\documentclass[sigconf]{acmart}

\usepackage{tabularx}
\usepackage{tablefootnote}
\usepackage{subcaption}
\usepackage{csquotes}

\AtBeginDocument{%
  }

\copyrightyear{2025}
\acmYear{2025}
\setcopyright{cc}
\setcctype{by}
\acmConference[Websci '25]{Proceedings of the 17th ACM Web Science Conference 2025}{May 20--24, 2025}{New Brunswick, NJ, USA}
\acmBooktitle{Proceedings of the 17th ACM Web Science Conference 2025 (Websci '25), May 20--24, 2025, New Brunswick, NJ, USA}
\acmDOI{10.1145/3717867.3717869}
\acmISBN{979-8-4007-1483-2/2025/05}

\graphicspath{{./images/}}

\begin{document}

\title{Multilingualism, Transnationality, and K-pop \\in the Online \#StopAsianHate Movement}


\author{Tessa Masis \emph{(they/them)}}
\affiliation{%
  \institution{University of Massachusetts Amherst}
  \city{Amherst}
  \state{MA}
  \country{USA}
}
\email{tmasis@cs.umass.edu}

\author{Zhangqi Duan}
\affiliation{%
  \institution{University of Massachusetts Amherst}
  \city{Amherst}
  \state{MA}
  \country{USA}}

\author{Weiai Wayne Xu}
\affiliation{%
  \institution{University of Massachusetts Amherst}
  \city{Amherst}
  \state{MA}
  \country{USA}}

\author{Ethan Zuckerman}
\affiliation{%
  \institution{University of Massachusetts Amherst}
  \city{Amherst}
  \state{MA}
  \country{USA}}

\author{Jane Yeahin Pyo}
\affiliation{%
  \institution{University of Massachusetts Amherst}
  \city{Amherst}
  \state{MA}
  \country{USA}}

\author{Brendan O'Connor}
\affiliation{%
  \institution{University of Massachusetts Amherst}
  \city{Amherst}
  \state{MA}
  \country{USA}}
\email{brenocon@cs.umass.edu}

\renewcommand{\shortauthors}{Masis et al.}

\begin{abstract}
  The \#StopAsianHate (SAH) movement is a broad social movement against violence targeting Asians and Asian Americans, beginning in 2021 in response to racial discrimination related to COVID-19 and sparking worldwide conversation about anti-Asian hate. However, research on the online SAH movement has focused 
  on English-speaking participants so the spread of the movement outside of the United States is largely unknown. In addition, there have been no long-term studies of SAH so the extent to which it has been successfully sustained over time is not well understood. 
  We present an analysis of 6.5 million "\#StopAsianHate" tweets from 2.2 million users all over the globe and spanning 60 different languages, constituting the first study of the non-English and transnational component of the online SAH movement. Using a combination of topic modeling, user modeling, and hand annotation, we identify and characterize the dominant discussions and users participating in the movement and draw comparisons of English versus non-English topics and users. 
  We discover clear differences in events driving topics, where spikes in English tweets are driven by violent crimes in the US but spikes in non-English tweets are driven by transnational incidents of anti-Asian sentiment towards symbolic representatives of Asian nations. We also find that global K-pop fans were quick to adopt the SAH movement and, in fact, sustained it for longer than any other user group. 
  Our work contributes to understanding the transnationality and evolution of the SAH movement, and more generally to exploring upward scale shift and public attention in large-scale multilingual online activism. 
\end{abstract}




\begin{teaserfigure}
  \includegraphics[width=\textwidth]{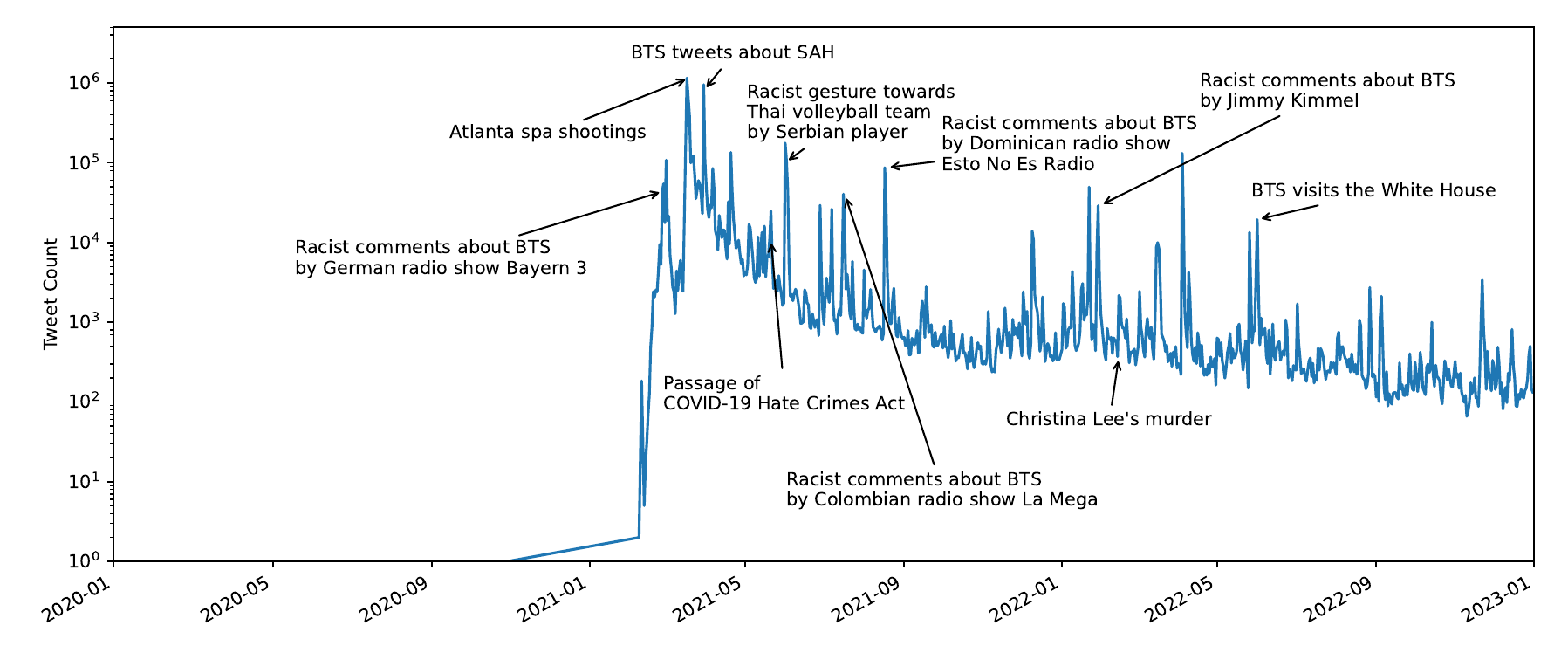}
  \caption{Daily tweet count of all tweets containing "\#StopAsianHate"; important events have been annotated. 
  }
  \label{fig:tweets-time}
  \Description[A graph with year and month on the x-axis, ranging from 2020-01 to 2023-01, and log-scale tweet count on the y-axis, ranging from $10^0$ to $10^7$. A blue line peaks around 2021-05 with $10^6$ tweets.]{A graph with year and month on the x-axis, ranging from 2020-01 to 2023-01, and log-scale tweet count on the y-axis, ranging from $10^0$ to $10^7$. A blue line peaks around 2021-05 with $10^6$ tweets, after which it hovers between $10^3$ and $10^5$ tweets. Annotations of important events, such as "BTS tweets about SAH" and "Passage of COVID-19 Hate Crimes Act", are connected with arrows to certain parts of the blue line.}
\end{teaserfigure}

\maketitle

\section{Introduction}

Since the outbreak of COVID-19, Asian Americans have faced increased discrimination and violence \cite{pew2022, committee23, staatus23}. On March 16 2021, the murder of six Asian women at three spas in Atlanta, GA, USA started a wave of rallies and protests across the United States and internationally. Many people spoke out against the rise of hate crimes against Asian people by using the \#StopAsianHate (SAH) hashtag on social media platforms like Twitter, now known as X. Despite the global attention of the movement, our review of the literature found no studies on SAH that have examined non-English tweets or corresponding users. 
Very few studies have examined users participating in the movement at all, and then only the most active users involved, and none have examined SAH data from after 2021.
We conduct a mixed methods analysis of all SAH tweets and corresponding user bios from 2020 to 2022, and examine: \textbf{RQ1)} how topics of discussion and users involved differ between English and non-English subsets, and \textbf{RQ2)} to what extent the online movement has been sustained over the years. Our data is the largest SAH dataset to date and our use of state-of-the-art natural language processing (NLP) methods allows us to analyze all 6.5 million tweets and 2.2 million user bios available.

Additionally, integrating information from multiple types of media can help us better understand social media's role in information spread and if it is leading or following the agenda set by traditional news sources \cite{zuckerman2021, freelon2018quantifying}. 
We calculate the cross-correlation between social media and mainstream media news time series to determine whether SAH tweets peak at the same time as mainstream news. 

More comprehensive analysis of the SAH movement can help inform racial justice activists about the reach and relevance of the movement and can contribute to our understanding of online activism more broadly. Unlike traditional activism, online social activism is not reliant on formal organizations to mobilize resources and public participation, and is instead able to function as decentralized, fluid in identities and goals, and geographically distributed \cite{bennett2012logic, vicari2017twitter}. Through our analysis, we hope to more clearly understand the transnationality of SAH and more generally how online movements can shift to and be sustained in transnational contexts.

Our work makes the following contributions: 
\begin{itemize}
    \item We release the first multilingual and the largest SAH dataset to date, which includes automatic translations of non-English tweets and user bios ($\frac{1}{3}$ of tweets and $\frac{1}{4}$ of users).\footnote{\url{https://github.com/slanglab/stopasianhate_data}}
    \item We model English and non-English tweets and users by clustering embedding representations of tweets and user bios, and contextualize the tweets in the broader media ecosystem by comparing with mainstream media news.
    \item We find that the SAH movement was evoked for different anti-Asian incidents in the United States versus globally: violent crime against ordinary individuals spurred tweet spikes in the US while elsewhere spikes were driven by hateful speech/gestures towards symbolic representatives of Asian nations, revealing a shift in frame needed to maintain relevance in a transnational and non-American context. 
    \item We identify global K-pop fans as the largest user group supporting the online SAH movement. 
    They were early adopters and the longest supporters of the movement, successfully sustaining it across geographic boundaries. 
\end{itemize}

\begin{figure*}[t]
  \centering
  \includegraphics[width=.7\textwidth]{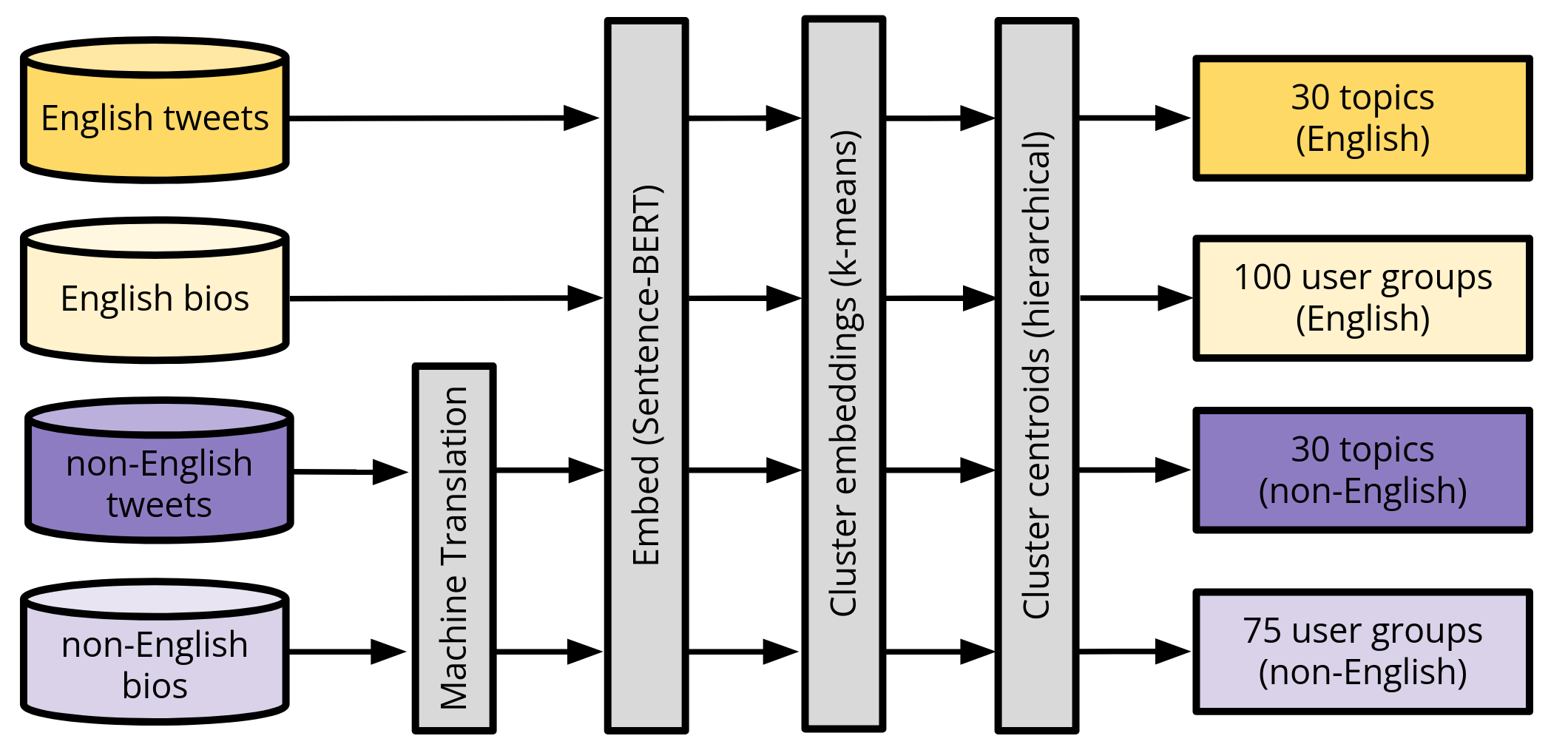}
  \caption{Pipeline for initial automatic clustering step in topic and user modeling. }
  \label{fig:method}
\end{figure*}

\section{Related Work} 

Our work is situated broadly within the space studying online or digital activism \cite{joyce2010digital} and is aligned with prior work studying other online movements and how they evolved over time, such as the BLM movement \cite{freelon2016beyond, giorgi2022twitter, le2022blm}. 
Although there have been quantitative studies of general anti-Asian discourse online \cite{jang2023anti, lakhanpal2022sinophobia, lin2022multiplex}, that body of work tends to center on instances of hate/toxic speech whereas our data from the SAH hashtag is a more Asian-sympathetic perspective with mostly people speaking out against anti-Asian hate crimes or sentiment. 

Initial studies of the online SAH movement explored salient topics in English/US-based SAH tweets, identifying topics related to the main concerns of the movement (e.g. history of anti-Asian hate \cite{cao2022stopasianhate}, recent anti-Asian hate crimes, white racism, and COVID-19 stigma \cite{guo2022blacklivesmatter, tong2022people}) as well as topics directly related to the Asian American and Pacific Islander (AAPI) community (e.g. supporting the AAPI community \cite{lee2021questing, chae2022stopasianhate} and wanting to increase the visibility of the AAPI community \cite{cao2022stopasianhate}). This body of work also identified Asian American influencers as vital drivers in promoting the movement \cite{fan2021stopasianhate, lyu2021understanding, wiguna2022communication} and saw that SAH was supported more in US states with higher rates of racially-motivated hate crimes \cite{lyu2021understanding}. 

A few papers have investigated intersectional questions such as the lack of significant relationship with or discussion of the \#BlackLivesMatter (BLM) movement in the SAH hashtag \cite{guo2022blacklivesmatter}, suggesting that the tagged interactions of online activism could create boundaries between imagined communities \cite{kim2023}. 
In our work, we investigate the involvement of imagined communities separated by geographical distance and language barriers, and how these barriers may be bridged during the formation of a transnational movement. 
To our knowledge, only one study has previously noted the transnationality of the SAH movement, examining how English tweets increased in volume and shifted from local to global vocabulary (e.g. "Atlanta", "US" to "world", "worldwide")
after BTS's tweet in support of SAH \cite{regla-vargas_alvero_kao_2023}. We seek to significantly build on this analysis by examining multilingual and global SAH tweets and the users who posted them. 

Previous work on SAH has led to findings about online activism more broadly. Research has explored the effects of moral and issue-based framing on SAH tweet engagement \cite{wang2023moral}, confirmed how online activity and offline events reinforce one another \cite{tong2022people}, and informed theoretical frameworks of social justice practices used by hashtag activists \cite{xie2023praxis}. 
In the present study, we seek to give insights into theories of upward scale shift related to transnational activism and theories of public attention related to sustaining social movements.

Notably, methods in the papers cited above studying SAH have typically involved either manual qualitative analysis of a small dataset (e.g. 1,000 tweets) 
or used computational methods such as Latent Dirichlet Allocation, the Structural Topic Model, network analysis, or hashtag frequency to analyze larger corpora of 0.1 to 5 million tweets.
The majority of the work has been short-term, focusing on a period of a few weeks or months, and only analyzed posts written in English and/or posted in the US. 
In contrast, we use state-of-the-art NLP methods to analyze a multilingual dataset of 6.5 million tweets posted globally over the span of 3 years, thereby expanding the scope of our results. 

\section{Data}

We collected 6,474,050 SAH tweets posted by 2,193,914 users in the years 2020 to 2022 (see Figure \ref{fig:tweets-time}). 
The English subset of our data (i.e. tweets identified as English) has 3,365,254 tweets posted by 1,367,658 users and the non-English subset (i.e. tweets identified as one of 59 non-English natural languages in the dataset) has 1,998,962 tweets posted by 577,628 users. 
Tweets and user bios in the non-English subset were automatically translated into English. 

\subsection{Data Collection}
The Twitter Academic Research Product Track V2 API was queried in early March 2023 using the R package \textit{academictwitteR} \cite{BarrieHo2021} to collect all tweets posted from January 1, 2020 to December 31, 2022 containing the string "\#StopAsianHate". The query is not case-sensitive and returns all historical tweets from the specified time period as long as the tweets were not removed by their authors or by content moderation before data collection.\footnote{This span includes 1012 days 
since the first mention of \#StopAsianHate on March 24, 2020; we are missing data from four days (2021-03-15, 2021-03-16, 2021-12-31, 2022-05-31) due to errors in the data collection process. While unfortunate, these do not impact our findings focusing on long-term trends in SAH.} Our dataset is similar in size to the \#BlueLivesMatter and \#StopTheSteal movements but an order of magnitude smaller than BLM \cite{giorgi2022twitter, chen2023comparing}.

Each tweet also contains metadata, including whether the tweet was a retweet, reply, or quote; how many times the tweet has been retweeted, replied to, or quoted; user handle; user description (also known as bio); and user location (free text field). All tweet and user metadata is from the time of data collection. 

Tweet metadata also includes automatically detected language of the tweet, as detected by Twitter's propriety language identification (LID) model. Although we do not have direct access to it, recent research suggests that, after 2016, it performs comparably with other tools used for LID of tweets \cite{alshaabi2021growing}. And, as we mention below, our analysis excludes ambiguous tweets for which no language classification could be made by Twitter's LID model.

\subsection{Data Processing}
There are 63 total language codes in the dataset which denote the language of the tweet as identified by Twitter. For each tweet and corresponding user bio identified as a non-English language, the Google Translate API was used to automatically translate the tweet and bio into English. Three of the language codes correspond to tweets containing only media links, hashtags, or emojis or to tweets which Twitter could not classify as a language: \emph{qme}, \emph{qht}, and \emph{und}. Tweets identified as such were not translated. We include in the dataset the original tweet text as well as a processed version with usernames and URLs removed, which we used for our analyses.

\section{Methodology}
\subsection{Topic and User Modeling}
\label{topic-modeling}
Due to the large size of the data, our overall methodology for both topic and user modeling is to conduct a thematic analysis where the initial coding iteration is done by clustering tweets/user bios embedded by a pretrained language model (Figure \ref{fig:method}) and the second iteration is done by two co-authors who manually label the clusters and group similar labels together into themes (Table \ref{table:english-topic-labels}). 
Our use of clustering follows from the consensus that clustering embeddings with k-means has better performance than LDA topic modeling for tweets \cite{demszky2019analyzing}, and clustering contextualized embeddings has comparable or better performance than LDA \cite{thompson2020topic, sia2020tired} or neural topic modeling \cite{zhang2022neural} for traditional documents. Clustering embeddings is also used in popular packages such as BERTopic \cite{grootendorst2022bertopic} and has been used successfully for modeling user groups \cite{xu2022unmasking}.
Our manual labeling follows the approach known as consensus coding, known to be a rigorous method for collaborative qualitative analysis \cite{richards2018practical}.

For topic modeling, we use the 200,247 English and 98,896 non-English tweets which were novel (i.e. not retweets, replies, or quote-tweets) and not duplicates (i.e. tweets that had the exact same text as another, indicating `spamming'). For user modeling, we use the 1,200,389 user bios in the English subset and 473,046 user bios in the non-English subset which were not empty, including all versions from every user.\footnote{User metadata is from the time of data collection and our data collection process took place over several weeks, so users who posted multiple times in the hashtag and changed their bio in the interim have multiple versions of bios in our dataset. This applies to less than 20,000 users in the dataset.} 

\textbf{Clustering.}
We use the same pipeline for both topic and user modeling, where the only difference is whether the input text is a tweet or user bio. First, the tweet/user bio is embedded with the all-MiniLM-L6-v2 model from Sentence-BERT \cite{reimers-2019-sentence-bert}, following the previously mentioned work that uses BERTopic-style topic modeling.
For tweets and bios in the non-English subset, the English automatic translations were embedded. Next, the embeddings are L2 normalized then clustered with the k-means implementation from \textit{sklearn} \cite{scikit-learn}.
An appropriate value for k was chosen by starting with k=5 and progressively trying higher values of k, manually examining the tweets/user bios nearest to the cluster centroids to select k when, instead of only producing clusters which were completely distinct from one another, there were some clusters that appeared identical to each other.
This criteria helped us determine a value k that would result in reasonably distinct topics/user groups -- a lower k would collapse dissimilar clusters together while a higher k would result in more identical clusters. 
Based on this, our final analysis used k=30 for English and non-English topic models, k=100 for the English user model, and k=75 for the non-English user model.

\begin{table}[h]
    \centering
    \begin{tabular}{l l}
    \hline
    \textbf{Initial Topic Label} & \textbf{Final Topic Label}\\
    \hline\hline
    Celebrate AAPI Heritage Month & AAPI Heritage Month\\
    Solidarity w/AAPI community & Solidarity w/AAPI community\\
    Public figures speak about SAH & Advocating action \& awareness\\
    Rallies, protests, activism & Advocating action \& awareness\\
    Hate crime legal reform & Advocating action \& awareness\\
    Donate to SAH & Advocating action \& awareness\\
    Raising awareness & Advocating action \& awareness\\
    SAH-related streaming & Advocating action \& awareness\\
    AAPI hate crimes & Anti-Asian hate crimes\\
    Atlanta spa shootings & Anti-Asian hate crimes\\
    NYC-related AAPI hate crime & Anti-Asian hate crimes\\
    SAH statement & SAH statement\\
    SAH-related merchandise & Commodification of SAH\\
    Yan Limeng's COVID misinfo & COVID-related misinfo/stigma\\
    COVID-19 stigma & COVID-related misinfo/stigma\\
    Connection with BLM & Connection with BLM\\
    Racism isn't comedy; BTS & BTS/ARMY\\
    Condemnation of racism; BTS & BTS/ARMY\\
    Apologize to BTS & BTS/ARMY\\
    \hline
    Thai-Serbia volleyball game & Advocating action \& awareness\\
    Condemning anti-Asian hate & Advocating action \& awareness\\
    AAPI hate crimes & Anti-Asian hate crimes\\
    SAH statement & SAH statement\\
    Yan Limeng's COVID misinfo & COVID-related misinfo/stigma\\
    BTS & BTS/ARMY\\
    K-pop concert & BTS/ARMY\\
    Racism isn't comedy; BTS & BTS/ARMY\\
    Billboard is racist; BTS & BTS/ARMY\\
    Apologize to BTS & BTS/ARMY\\
    Support for BTS & BTS/ARMY\\
    Unidentifiable & Other/Unknown\\
    \hline
    \end{tabular}
    \caption{For each automatically derived cluster, two coders conferred and decided on an initial topic label. Later, they merged similar initial labels to create final topic labels. The top half of the table refers to labels for English tweet clusters; the bottom half to labels for non-English tweet clusters. }
    \label{table:english-topic-labels}
\end{table}

Because we often had a relatively high value for k, it was useful to order the clusters in a semantically meaningful way. We hierarchically clustered the k cluster centroids using the hierarchical clustering function from \textit{scipy} \cite{2020SciPy-NMeth}, ordering them by the left-to-right traversal of the leaves in the dendrogram. This significantly sped up the subsequent manual labeling of the clusters, since highly related clusters were near each other.

\textbf{Thematic analysis.}
Clusters were manually labeled and similar labels were grouped together into themes. Each cluster was represented in a spreadsheet with a range of descriptive statistics and examples, including number of tweets/bios in the cluster and the 10 tweets/bios closest to the centroid. (Some clusters included additional information; for example, non-English clusters included the top three languages and their proportion in the cluster. For complete information on how clusters were represented and files containing all cluster representations, see the released dataset and documentation.) Two coders, both co-authors on this paper, independently read the information and labeled each cluster. The coders were instructed that each label should be only a few words long and should encode the most salient topic or characteristic of the cluster.
Afterward, the two coders met, discussed their labels, and decided together on initial labels for each cluster. 
The coders then went through the initial labels and merged similar ones, until there were 10 total topic labels and 8 total user group labels, each corresponding to a set of automatically derived clusters (see Table \ref{table:english-topic-labels} for topic labels and Appendix \ref{sec:cluster-labels} for user group labels; see Appendix \ref{sec:tweet-bio-examples} for examples of tweets and user bios from select clusters).

\subsection{Comparing Twitter with Mainstream Media}
To investigate whether social media was leading or following mainstream media news, we collected counts over time of SAH-related mainstream news articles and computed their cross-correlation with SAH tweet volume over time. 
Cross-correlation, a measure of lagged similarity between two time series, indicates at which point the time series best match up. If the best match occurs, for example, when mainstream news is delayed by one day,
this indicates that the SAH hashtag is heavily used the day after news reports of an anti-Asian incident and is thus following mainstream media.

Media Cloud, an open data platform for storing, retrieving, and analyzing online news \cite{roberts2021media}, was used to collect the number of news articles posted per day with the keywords ``stopasianhate'' or ``anti-asian'' from January 1, 2020 to December 31, 2022. We collected the number of articles posted in the US National collection and the number of articles posted in any of 7 National collections of Asian countries (Thailand, Korea, Japan, Indonesia, Vietnam, Philippines, and China; these countries were chosen due to being the most frequent Asian languages in our dataset). From our SAH dataset, we calculated the number of novel tweets posted per day in the English subset and the number posted per day in the non-English subset. We thus have four time series: SAH-related articles published in US news, SAH-related articles published in Asian news, tweets posted in our English subset, and tweets posted in our non-English subset. 
We used \textit{statsmodels} \cite{seabold2010statsmodels}
to calculate cross-correlation functions between all pairings of the four time series.

\section{Results}

\subsection{English Subset}
\textbf{Topics.}
Many of the topics identified in the English subset (Table \ref{table:english-topics})
broadly align with and confirm those found in previous studies, including statements of solidarity with the AAPI community \cite{lyu2021understanding}, advocating action and awareness \cite{lee2021questing}, discourse on recent anti-Asian hate crimes \cite{guo2022blacklivesmatter, lyu2021understanding}, COVID-related misinformation or stigma \cite{guo2022blacklivesmatter, tong2022people}, and
the connection between SAH and BLM \cite{lyu2021understanding}. Our analysis also identifies novel topics, such as  tweets related to BTS  where fans -- called ARMY -- call out radio and TV show hosts for racist comments about BTS (15\% of tweets) and tweets related to AAPI Heritage Month (3\% of tweets).

Although there is not a substantial amount of discord within the majority of topics and most users in the SAH hashtag seem to agree on what they think deserves to be supported or condemned, there is noticeable division when it comes to the relationship between Asian and Black communities. Although there is broad support for BLM within the SAH movement, there are also those who oppose the movement, e.g. both \#blacklivesmatter and \#whitelivesmatter are frequent hashtags in the "Connection with BLM" cluster. There are also users expressing wishes that BLM was supported as widely as SAH or complaints that Black people aren’t supporting SAH the same way that Asians supported BLM (see Appendix \ref{sec:tweet-bio-examples}). Within the "Anti-Asian hate crimes" clusters, \#stopasianhateagainstblackpeople is a frequent hashtag, referring to anti-Black sentiment that can be prevalent in Asian American communities. We therefore see evidence of both anti-Black sentiment, from white and Asian communities, as well as dissatisfaction about lack of solidarity between Asian and Black communities. 
This finding supports prior work noting explicit interracial tensions in the SAH hashtag between Black and Asian communities \cite{lyu2021understanding}, as well as work pointing out the implicit lack of connection and solidarity between online BLM and SAH movements \cite{kim2023, guo2022blacklivesmatter, tong2022people}. 

\textbf{User groups.}
Descriptive statistics about the user groups identified can be seen in Table \ref{table:english-users}.\footnote{User group percentages should be interpreted as lower bounds, due to two reasons. First, user bios contain a curated subset of identity markers and do not encompass an entire user's identity, so there is almost certainly not enough information to accurately label every user. Indeed, 17\% of bios (those in the "Other/Unknown" user group) did not contain any useful identity information at all. Second, people often belong to multiple identity groups and many users had multiple identity markers in their bio, but each user in our dataset was only assigned to one user group label.}
Notably, we discover that the largest cohesive group of users posting in the SAH hashtag are K-pop fans. They participate more than average (2.9 tweets per user, versus the average of 2.5) and they make up at least 27\% of the users, more than double the combined number of all other similarly active user groups (users with political affiliations, identifying as activists, or identifying as East/Southeast Asian, who all posted more than average in the hashtag, make up only 12\% of users).

\begin{table}[t]
    \centering
    \begin{tabular}{l r}
    \hline
    \textbf{Topic} & \textbf{\%}\\
    \hline\hline
    AAPI Heritage Month & 2.9\\
    Solidarity with AAPI community & 24.2\\
    Advocating action and awareness & 21.1\\
    Anti-Asian hate crimes & 11.3\\
    SAH statement & 16.8\\
    Commodification of SAH & 2.7\\
    COVID-related misinfo/stigma & 3.2\\
    Connection with BLM & 3.1\\
    BTS/ARMY & 14.7\\
    \hline
    \end{tabular}
    \caption{Topics identified in the English subset, and their percentage out of all novel English tweets. }
    \label{table:english-topics}
\end{table}

\begin{table}[t]
    \centering
    \begin{tabular}{l r r r}
    \hline
    \textbf{User group} & \textbf{\%} & \textbf{\# twts} & \textbf{\% RT}\\
    \hline\hline
    K-pop fan & 26.8 & 2.9 & 94.7 \\
    Political affiliation in bio & 3.1 & 3.6 & 95.0\\
    Activist & 7.8 & 2.7 & 87.4\\
    East/Southeast Asian identity & 1.8 & 3.3 & 87.5\\
    National identity (non-Asian) & 3.5 & 1.9 & 95.1\\
    Job & 14.4 & 2.4 & 86.7\\
    Other identity marker\tablefootnote{User bios which only have pronouns, astrological sign, age, location, quote, familial role, hobbies, or religious identity. } & 25.2 & 2.3 & 93.5\\
    Other/Unknown\tablefootnote{Contained some mix of symbols, numbers, and/or languages which were not interpretable to the coders.} & 17.4 & 2.3 & 94.4\\
    \hline
    Average across all user groups & & 2.5 & 92.4\\
    \hline
    \end{tabular}
    \caption{User groups identified in the English subset, their percentage out of all English user bios, average number of tweets per user, and percentage of retweets out of all tweets posted by user group. }
    \label{table:english-users}
\end{table}

The high participation we see from East/Southeast Asian users and activist users is consistent with findings from a previous study examining users involved in SAH \cite{lyu2021understanding}.
East/Southeast Asian users tweet the second most out of all the user groups (3.3 tweets per user, versus the average of 2.5) and write more novel tweets than average (88\% retweets, versus the average of 92\%), and activist users similarly tweet and write more novel tweets than average (2.7 tweets per user; 87\% retweets). 

\subsection{Non-English Subset}
\textbf{Topics.} We discover that many of the conversations in the non-English tweets (Table 
\ref{table:noneng-topics})
are driven by incidents happening in local, non-US contexts and discussed in the relevant language. For example, within the "Advocating action and awareness" topic, one cluster is composed of Thai users expressing outrage at a Serbian player's racist gesture during a Thai-Serbian volleyball game (2\% of tweets; 78\% of tweets in this cluster are in Thai). 
Within the "COVID-related misinfo/stigma" topic, one cluster contains Chinese users denouncing Yan Limeng, a public figure who spread misinformation about COVID-19 (1\% of tweets; 77\% of this cluster is in Mandarin).
And within the "BTS/ARMY" topic, there are multiple clusters containing K-pop fans calling out Colombian and Dominican radio show hosts, with the majority of these tweets in Spanish. 
On the whole, the "BTS/ARMY" topic is over three times as large as in the English subset (51\% of tweets versus 15\%) with tweets covering a wider catalog of BTS-related incidents, including racist jokes by radio show hosts in Germany and Colombia. 

These topics reveal how the SAH movement is not confined to the US, or even to Western countries. While there has been some recognition of anti-Asian racism and SAH outside of the US \cite{Chen2021}, the focus has been on Western and English-speaking countries like Canada, the UK, and Australia, and researchers have not examined the online SAH movement outside of these countries. We see here that there are substantial conversations in SAH happening in Asian and Latin American countries, in the languages spoken by the people living there, none of which have been covered in prior work.

\textbf{User groups.}  
Similar to our finding of the English subset, the largest cohesive group of users are K-pop fans (Table \ref{table:noneng-users}). Here, they constitute more than \emph{all} of the other user groups combined, making up more than 62\% of non-English users (compared to 27\% of English users). We also see that non-English activist users, similar to English ones, write more novel tweets than average (89\% retweets, versus the average of 94\%). 

We examined the 10 most frequent words in users' `location' field for all 75 user clusters, using automatic translation for words in non-Latin scripts and ignoring words that were not geographic locations (e.g. pronouns, band names). We observed that many of the K-pop fans are from Thailand, Indonesia, Vietnam, Mexico, Argentina, Brazil, Colombia, and Peru. These results are more suggestive than conclusive -- previous work has shown that, although more than 40\% of Twitter users include recognizable locations in their `location' field, different user groups are more likely to actually use this field or to geotag their tweets (e.g. 3\% of Korean-speaking users use geotags, compared to 42\% of Indonesian-speaking users) \cite{huang2019large}. However, combined with the fact that all the users in the non-English "East/Southeast Asian identity" group identified as Thai and all the users in the non-English "National identity (non-Asian)" group identified as Latin American, as well as with our findings of topics specific to Asian and Latin American countries, it suggests substantial participation from Southeast Asian and Latin American countries within the online SAH movement.

\subsection{Comparison with Mainstream Media}

\begin{figure}[t]
  \centering
  \begin{subfigure}[h]{.48\textwidth}
    \centering
    \includegraphics[width=\textwidth]{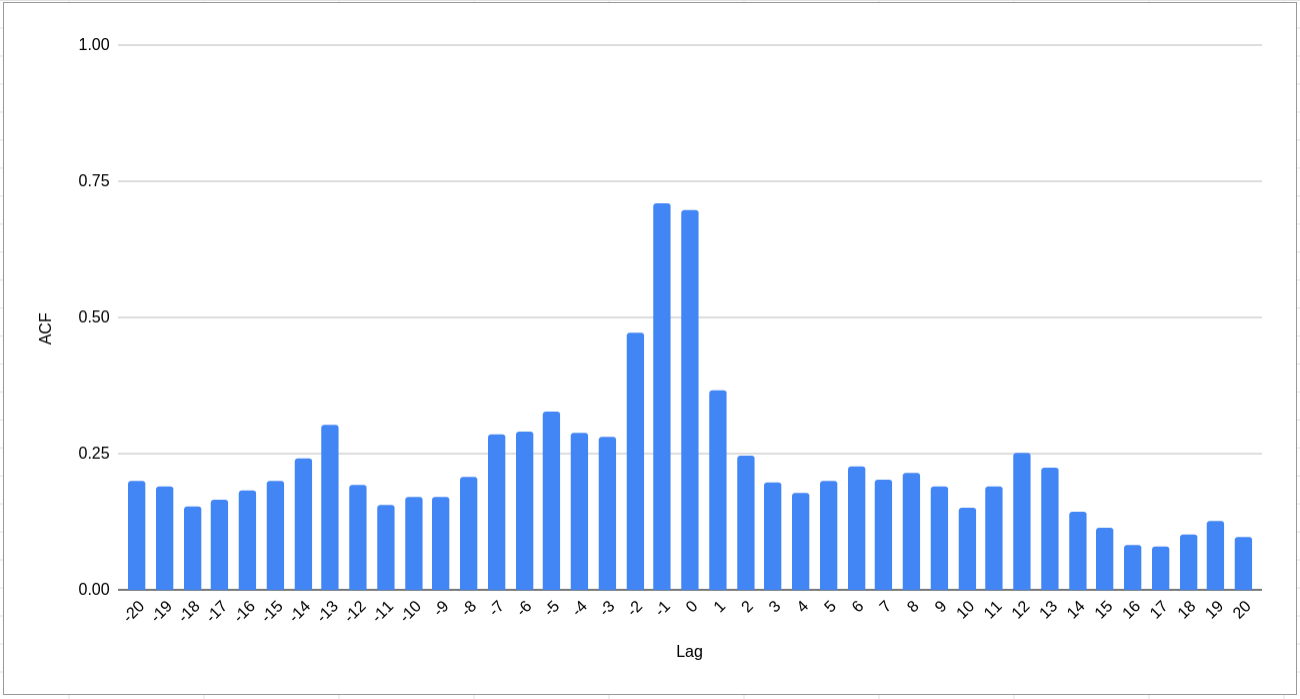}
    \caption{Cross-correlation of English Twitter \& US news}
  \end{subfigure}
  \begin{subfigure}[h]{.48\textwidth}
    \centering
    \includegraphics[width=\textwidth]{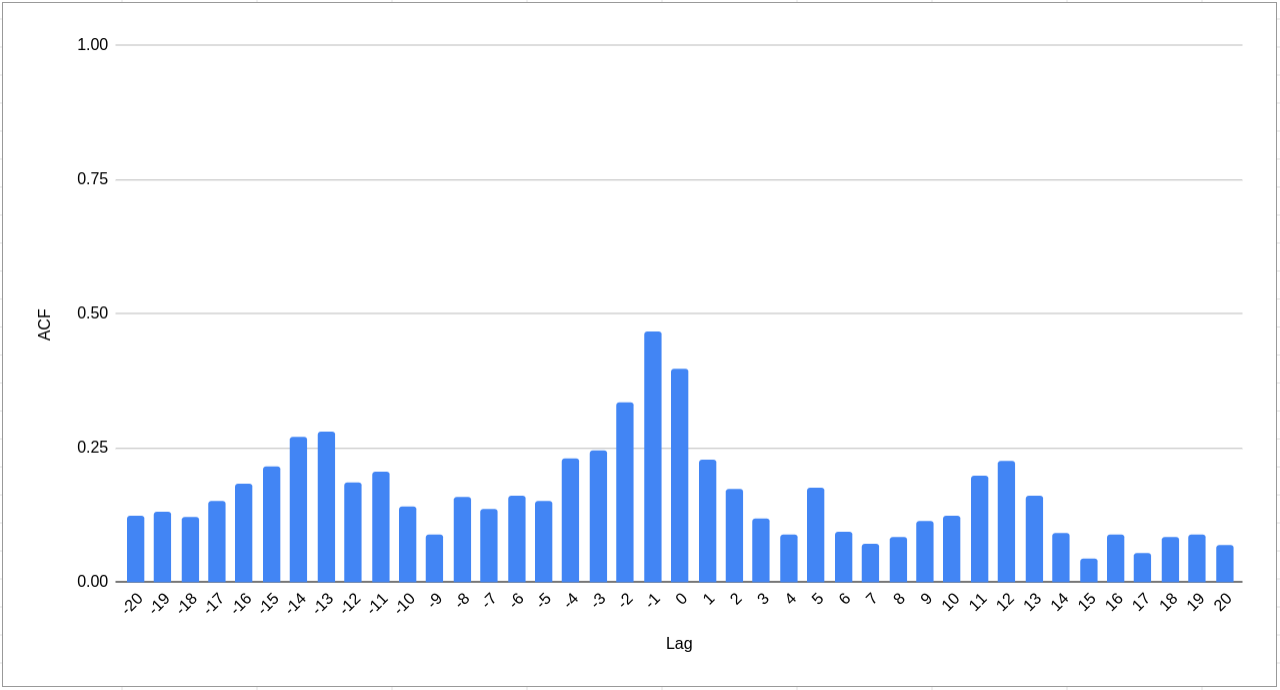}
    \caption{Cross-correlation of non-English Twitter \& Asian news}
  \end{subfigure}
  \caption{Peak cross-correlation for both (a) English Twitter \& US news and (b) non-English Twitter \& Asian news occurs when lag is -1, indicating that social media is not lagging behind the press but leading it.}
  \label{fig:cross-corr}
\end{figure}

We see that mainstream news lags one day behind the Twitter time series, with the peak cross-correlation between a Twitter and a mainstream news time series ($r = .71$ for English Twitter and US news, $r = .47$ for non-English Twitter and Asian news) occurring when lag is -1 (Figure \ref{fig:cross-corr}). 
This indicates that social media is leading the press -- a surprising finding, given that the most common pattern for online social movements is to react to and amplify the agenda set by traditional news sources (other notable exceptions where social media may lead the press include BLM \cite{freelon2016beyond} and \#MeToo). 
Less surprisingly, the time series are globally aligned within media types. The peak cross-correlation between US news and Asian news ($r = .77$) occurs when lag is 0, suggesting that the two time series are aligned. We see the same pattern with the English and non-English Twitter time series, where its peak cross-correlation ($r = .88$) is also when lag is 0.

\begin{table}[t]
    \centering
    \begin{tabular}{l r}
    \hline
    \textbf{Topic} & \textbf{\%} \\
    \hline\hline
    Advocating action and awareness & 12.1 \\
    Anti-Asian hate crimes & 9.1\\
    SAH statement & 21.5\\
    COVID-related misinfo/stigma & 1.2 \\
    BTS/ARMY & 50.7\\
    Other/Unknown & 5.4\\
    \hline
    \end{tabular}
    \caption{Topics identified in the non-English subset, and  their percentage out of all novel non-English tweets. }
    \label{table:noneng-topics}
\end{table}

\begin{table}[t]
    \centering
    \begin{tabular}{l r r r}
    \hline
    \textbf{User group} & \textbf{\%} & \textbf{\# twts} & \textbf{\% RT}\\
    \hline\hline
    K-pop fan & 62.1 & 4.7 & 94.7\\
    Activist & 1.6 & 3.5 & 89.3\\
    East/Southeast Asian identity & 0.5 & 4.3 & 94.3\\
    National identity (non-Asian) & 1.2 & 3.1 & 91.7 \\
    Job & 4.3 & 3.2 & 89.6\\
    Other identity marker & 23.8 & 3.8 & 94.9\\
    Other/Unknown & 6.6 & 4.0 & 93.5\\
    \hline
    Average across all user groups & & 4.3 & 94.3\\
    \hline
    \end{tabular}
    \caption{User groups identified in the non-English subset, their percentage out of all non-English user bios, average number of tweets per user, and percentage of retweets out of all tweets posted by user group.  }
    \label{table:noneng-users}
\end{table}

While the majority of ground events (e.g. Atlanta spa shootings, passage of the COVID-19 Hate Crimes Act, the Thai-Serbia volleyball game) are covered by both traditional news and Twitter, we identified several events which were only discussed by one media source, via a manual examination of peaks in the US news and English tweets time series. The first identified event is Christina Lee's murder: while it was indeed covered by traditional news sources, there is no peak in tweets at the relevant time (see Figure \ref{fig:tweets-time}). This finding is also supported by our topic analysis of the English subset of tweets, where we found very little discussion of her murder (see Table \ref{table:english-topic-labels}). The two other identified events are incidents of racist comments about BTS being made by radio show hosts in Colombia and the Dominican Republic. These events were discussed only on Twitter, with little news coverage.

\section{Discussion}

Our results indicate that a significant number of SAH conversations occurred in languages other than English and about anti-Asian incidents outside of the US. Importantly, there is also a distinction in both the type and targets of events which drive spikes in the SAH hashtag in the US versus globally, highlighting necessary changes that occur when a movement scale shifts from the national level to the transnational. We discuss this here and additionally discuss the involvement of global K-pop fans, who are the largest cohesive user group supporting the online SAH movement and were able to bridge national divides to collectively mobilize around anti-Asian incidents occurring in non-US contexts. Their sustained activity in a transnational social movement offers insights into effective online mobilization.

\subsection{SAH in the US versus Globally}

In comparing topics in English versus non-English tweets, we find that violent crime frequently drove conversation in English tweets while it was non-criminal racist incidents which drove conversation in non-English tweets. 
In English tweets, there is more discussion in general about violent crimes -- 11\% of English tweets are in the "Anti-Asian hate crimes" topic, versus only 9\% of non-English tweets, and an additional 3\% of English tweets discuss SAH's connection with BLM, a movement centering on racially-motivated violence. As corroborated by previous work \cite{fan2021stopasianhate, guo2022blacklivesmatter, tong2022people} and visible in Figure \ref{fig:tweets-time}, events driving English conversations often concern anti-Asian violence, such as the Atlanta spa shootings, news of other recent hate crimes, and the passage of the COVID-19 Hate Crimes Act.
In contrast, we find that events driving conversation in non-English tweets were primarily concerned with non-criminal incidents such as racist comments, jokes, and gestures, rather than instances of anti-Asian violence. As noted earlier, more than three times as many non-English tweets are related to ARMY defending BTS against racist comments (51\% vs 15\%) and 2\% of all non-English tweets are concerned with an incident where a Serbian volleyball player made a racist gesture at the Thai volleyball team. These non-criminal incidents drove large spikes in tweet volume (Figure \ref{fig:tweets-time}).  
 
Additionally, while the targets of anti-Asian incidents driving conversation in the US were typically everyday individuals, the events driving non-English tweets are primarily concerned with celebrities or public figures. 
In a transnational context, these public figures can function as representatives of Asian nations. 
The Thai women's national volleyball team expressly represents their country in international games and is an especial source of national pride for Thai people as one of the top teams in the world \cite{Duankloy_2022}. 
Due to their exceptional worldwide success, BTS has ties to South Korean national identity, and have spoken twice at the United Nations as representatives of South Korea and visited the White House for AAPI Heritage Month \cite{Vigdor_2021, Regan_Dailey_2022}. 
Both BTS and the Thai women's volleyball team represent their respective countries on global stages, so when they are recipients of anti-Asian racism it can also be interpreted as their home country which is the target. 

We hypothesize that these differences in type and target of anti-Asian incidents in a US versus transnational setting -- violent crime against ordinary individuals versus racist sentiment against symbolic representatives of Asian nations -- are indicative of "scale shifting", or a change in the level of activism which can include new actors or objects and broadened claims \cite{mcadam2003dynamics, tarrow2005new}. 
The issue transcended the narrative in which SAH was originally linked regarding racially-motivated violence in the US, and broadened from individual-level incidents to nation-level ones and from violent crimes to incidents of general anti-Asian sentiment. 

We also note differences in the online SAH movement when comparing individual non-English languages. For example, 66\% of non-English tweets are in Thai, four times more than any other language, and 29\% of novel non-English tweets are in Spanish, the most of any language. Some of this can be explained by differing Twitter cultures -- Spanish was the third top natural language by volume in 2020 and Thai was the top language in terms of retweeting ratio \cite{alshaabi2021growing}. In other words, there are more Spanish tweets and Thai retweets than other languages so they have a greater representation in our dataset. In light of these statistics, it is the large proportion of Vietnamese tweets in our dataset that stands out as surprising: despite not being in the top 30 languages on Twitter for the past ten years, it is the 11th most frequent language in our non-English subset. After examining the Vietnamese tweets in our dataset more closely, their presence in SAH seems to be due mostly to V-ARMY (the Vietnamese branch of ARMY) with primarily tweets about BTS.

Another explanation for why the online SAH movement looks different across languages may be because of the political climates in their respective countries and the role that social media serves when there is limited freedom of the press. There are numerous instances of communities using social media for activism and political change by using it as a tool for creating and distributing content without government control \cite{kyriakopoulou2011authoritarian}. Notably, many of the users in our non-English subset are based in Thailand, Vietnam, and Mexico, which are known to be countries with limited freedom.\footnote{freedomhouse.org/explore-the-map} In these countries, social media users may be more used to using social media as a tool with which to spearhead social justice movements on, thereby making it easier for them to participate in the SAH movement.

\subsection{K-pop Fans and Sustained Transnational Activity}

The largest group of users participating in the online SAH movement are K-pop fans, making up more than one fourth of users in our English subset and more than half of users in our non-English subset. They mobilized around incidents occurring in countries across the world (e.g. Colombia, the Dominican Republic, the US, Germany) and shared information across languages (e.g. the clusters frequently contained high percentages of tweets in multiple languages). 
In our comparison of Twitter versus mainstream media news, we also saw that BTS-related incidents are the only events which are only on Twitter and not covered by traditional news sources. It is therefore necessary to recognize the importance of BTS and K-pop fans in general in supporting the SAH movement. 

K-pop fans were some of the earliest adopters of the hashtag, using it before BTS had even endorsed the movement, and some of the longest supporters, mobilizing for BTS-related events well into 2022 (see Figure \ref{fig:tweets-time}). 
In contrast, non-BTS related topics were receiving much less attention by 2022, with the last non-BTS tweet spike in June 2021 even though there were still anti-Asian hate crimes occurring. For example, Christina Lee's murder occurred in February 2022 but received remarkably little attention in our dataset -- as we saw in our comparison of Twitter versus mainstream media news, while Christina Lee's murder was covered by traditional news it was hardly talked about on Twitter, with no visible peaks in that time range and no event-specific topic clusters identified. We hypothesize that this is related to Downs's theory of "issue-attention cycles", where a social problem leaps into prominence for a short time before fading from public attention, even as the problem remains relevant and unsolved \cite{downs1972up}. This sort of attention cycle was shown to be apparent in BLM, where there was a particular period of time when hate crimes were more open to being reported on or discussed \cite{zuckerman2019whose}. The Atlanta spa shooting was a key event in SAH that led to increased public discussion and awareness of anti-Asian hate crimes, but by Christina Lee's murder 11 months later -- another hate crime involving the killing of Asian women -- this period of increased public attention had ended and her death, along with other instances of violent hate crimes, received the same minimal attention as they would have before the Atlanta shootings. 
This decline in attention may also be related to concerns about the sustainability of online activism without having core organizers to sustain the movement \cite{tufekci2017twitter} and due to the ephemerality of attention in the digital environment.

So why were K-pop fans less affected by the issue-attention cycle and able to continue sustained activity in the hashtag for longer than any other user group, effectively mobilizing for SAH across geographic and linguistic boundaries for more than 1.5 years? 
To answer this, we draw on related work examining how people use culture products as momentum to accomplish political goals \cite{jenkins2015cultural, brough2012fandom, kligler2016decreasing}. Fan groups not only forge collective identities around specific media but also create communication infrastructure, social networks, and fan rituals and rhetoric, all of which can be utilized to great effect when enacting activist agendas. In contrast, `Asians' and `Asian Americans' as a whole do not have dense social networks and, in fact, often find themselves divided by their diverse identities, experiences, and struggles \cite{kibria1998contested}. We hypothesize that the preexisting fan practices, transnational social networks, and collective identities of K-pop fans allowed them to engage in sustained collective action for longer than users connected solely by ambiguous pan-ethnic terms such as `Asian' and `Asian American'.

K-pop stans have been noted previously for their ability to effectively organize and mobilize in both cultural and civic spheres. 
Recent work has acknowledged the outsize role of BTS and ARMY on the SAH movement, investigating the impact of BTS's tweet supporting SAH -- the most retweeted tweet in 2021 -- on the volume of SAH tweets in the following weeks \cite{regla-vargas_alvero_kao_2023}. 
Previous studies have additionally examined the participation of BTS and their fans in the BLM movement, where BTS and ARMY each donated 1 million dollars to BLM and ARMY hijacked the \#whitelivesmatter and \#bluelivesmatter hashtags \cite{cho2022bts, park2021armed, lee2021make, kim2021k}. 
K-pop fans more generally have been acknowledged as a sociopolitical force for years, especially in Southeast Asian and Latin American countries \cite{han2017korean, andini2021exploring, chang2023parasocial}. 


\section{Conclusion}
This paper constitutes the largest dataset and the first multilingual analysis of the online SAH movement to date. Using state-of-the-art NLP methods to model tweets and users, we are able to put forward answers to key questions about the transnationality of the SAH movement and how it has evolved over time. Our study demonstrates the substantial participation in SAH of people speaking in languages other than English about anti-Asian incidents outside of the US, particularly from East Asia, Southeast Asia, and Latin America. While all previous studies of discourse in the online SAH movement have exclusively examined English tweets and discussion of the movement outside the US has only been concerned with English-speaking countries, our data makes clear that countries outside of the US are participating in SAH.

We find that there are distinct differences in the type and target of events which drive US versus global conversations around SAH. While the former focuses on violent hate crimes against ordinary individuals, the latter is more concerned with general anti-Asian sentiment towards national symbols. 
Additionally, we find that K-pop fans are the largest unified group of users participating in SAH, with a significant presence in sustaining the movement. We hope to have shed more light on the online SAH movement as well as on the phenomena of transnational, online, and fan activism. 

We acknowledge some limitations of this work. First, our analysis relies on automatic translations of non-English tweets, which may inhibit cross-lingual and/or cross-cultural understandings. 
Second, we do not include data or analysis of the online SAH movement on other social media platforms, such as YouTube or Reddit, which we leave to future work. 

As with any social media dataset, there are ethical concerns to consider when sharing the data. Most concerns are related to the possibility of identifying individuals within the dataset or identifying personal information about such individuals (i.e. political stance, location). These actions are possible if the individuals have at some point shared identifying information on their public account. Although the majority of the users participating in SAH are doing so with the intention of their statements being public, it is also true that most are doing so with the implicit guarantee of "security through obscurity" -- the majority of users are not public figures or influencers but regular people who rely on relative inconspicuousness to maintain a sense of anonymity. We acknowledge that by curating and releasing this dataset and making their data more easily findable, we are undermining their obscurity and thus their security. We also note that no individuals in the dataset have explicitly consented to having their data shared by the authors. 

Due to these concerns, we choose to release our dataset only for non-commercial academic or journalistic research purposes, in order to limit the possibility of the dataset being misused. We hope that future researchers will use this dataset for good faith investigations to enhance world knowledge and that it will allow them to more easily examine the involvement of international users in SAH, including analyses of their discourse, temporal shifts, or spatial trends. This dataset would also be valuable for comparisons with other online activism movements.

\begin{acks}
We would like to thank the anonymous reviewers for their helpful comments and feedback. 
This material is based upon work supported by National Science Foundation grants 49100421C0035, 1845576, and a Graduate Research Fellowship (1938059).  Any opinions, findings, and conclusions or recommendations expressed in this material are those of the authors and do not necessarily reflect the views of the National Science Foundation. This work was also supported by the UMass Amherst Data Science for the Common Good program (DS4CG, via the Center for Data Science).

\end{acks}

\bibliographystyle{ACM-Reference-Format}
\bibliography{references.bib}

\newpage
\appendix

\section{Example Tweets and User Bios}
\label{sec:tweet-bio-examples}

Included here are synthetic tweets and user bios, manually written by the first author, intended to be illustrative of 3 topics and 3 user groups while also protecting the privacy of the users. Tweets from the "Anti-Asian hate crimes" topic:
\begin{itemize}
    \item The shootings in Atlanta were grotesque, hateful, and yet another reminder that we must be vigilant in fighting against hatred. Violence against the Asian community must be stopped. We must commit to \#StopAsianHate
    \item there has been a staggering 1900\% increase in targeted hate crimes against asian americans. the old, young, and most vulnerable have been targeted assaulted and killed. this needs to STOP. \#StopAsianHate \#AsiansAreHuman
\end{itemize}

Tweets from the "COVID-related misinfo/stigma" topic:
\begin{itemize}
    \item \#StopAsianHate if you call covid-19 the Chinese Virus, then you part of the problem. Stop the hate yall... we trying to survive and stay healthy just like everyone else.
    \item The Asian community has been experiencing racism long before COVID, but with this virus there’s been too much of an increase in hate \& violence against the community. It's absolutely disgusting \#StopAsianHate \#StopAsianHateCrimes
\end{itemize}

Tweets from the "Connection with BLM" topic:
\begin{itemize}
    \item Let's start spreading \#StopWhiteAmericanBeingRacist as well. \#BlackLivesMatter \#StopAsianHate 
    \item I wish \#BlackLivesMatter was as widely supported as \#StopAsianHate, the racism is palpable and disgusting! stop racism!
\end{itemize}

User bios from the "K-pop fan" user group:
\begin{itemize}
    \item love you more than love \#BTSARMY
    \item GOT7FOREVER // NCT DREAM // *Wooyoung noticed me*
\end{itemize}

User bios from the "Political affiliation in bio" user group:
\begin{itemize}
    \item Proud Democrat, Liberal, Activist, Pro-Choice. I hate Nazis, Fascists, and Republicans \#VoteBlue \#Resist \#LoveIsLove
    \item Christian, 100\% Conservative, God believer, American, Trump Supporter, GoTrump2020 \#MAGA
\end{itemize}

User bios from the "East/Southeast Asian identity" user group:
\begin{itemize}
    \item Taiwanese American $\vert$ M.A., East Asian studies $\vert$ Associate Professor $\vert$ tweets are my own $\vert$ she/they
    \item avid traveler, Tokyo-based reporter, trilingual translator. 
\end{itemize}

\newpage
\section{User Group labels}
\label{sec:cluster-labels}

\begin{table}[h]
    \centering
    \begin{tabular}{l l}
    \hline
    \textbf{Initial User Group Label} & \textbf{Final User Group Label}\\
    \hline\hline
    K-pop fan & K-pop fan\\
    BTS fan & K-pop fan\\
    Political affiliation & Political affiliation in bio\\
    Supports BLM & Activist\\
    Feminist & Activist\\
    Community organization & Activist\\
    Democracy quote & Activist\\
    Queer identity & Activist\\
    Has a disability & Activist\\
    East/Southeast Asian identity & East/Southeast Asian identity\\
    Indonesian identity & East/Southeast Asian identity\\
    Japanese identity & National identity (non-Asian)\\
    Latin American identity & National identity (non-Asian)\\
    French identity & National identity (non-Asian)\\
    Journalist/Writer & Job\\
    Artist/Musician & Job\\
    Movie industry & Job\\
    Veteran & Job\\
    Educator & Job\\
    Technology job & Job\\
    Marketing & Job\\
    CEO & Job\\
    Public affairs & Job\\
    Doctor & Job\\
    Selling merchandise & Job\\
    Astrology sign & Other identity marker\\
    Female pronouns & Other identity marker\\
    Male pronouns & Other identity marker\\
    Age & Other identity marker\\
    Location & Other identity marker\\
    Family role & Other identity marker\\
    Religious identity & Other identity marker\\
    Positive emotional quote & Other identity marker\\
    Negative emotional quote & Other identity marker\\
    Personality trait & Other identity marker\\
    Entertainment interests & Other identity marker\\
    Anime fan & Other identity marker\\
    Gamer/Twitch user & Other identity marker\\
    Foodie & Other identity marker\\
    Likes animals & Other identity marker\\
    Twitter-user identity & Other identity marker\\
    Generic fan account & Other/Unknown\\
    Banned/inactive account & Other/Unknown\\
    Emoji/links & Other/Unknown\\
    Unidentifiable & Other/Unknown\\
    \hline
    \end{tabular}
    \caption{Initial and final user group labels for English and non-English user bios. }
    \label{table:eng-user-labels}
\end{table}

\end{document}